%% file: main.tex
\documentclass[10pt,twocolumn,letterpaper]{article}

\usepackage[pagenumbers]{cvpr} 

\input{preamble}

\definecolor{cvprblue}{rgb}{0.21,0.49,0.74}
\usepackage[pagebackref,breaklinks,colorlinks,citecolor=cvprblue]{hyperref}

\def\paperID{*****} 
\def\confName{CVPR}
\def\confYear{2024}

\title{Instance Temperature Knowledge Distillation}

\author{Zhengbo Zhang$^1$, Yuxi Zhou$^2$,Jia Gong$^1$, Jun Liu$^1$, Zhigang Tu$^2$ \\
\textit{$^1$Singapore University of Technology and Design}, \textit{$^2$Wuhan University}
\\
{\tt\small zhengbo\_zhang@mymail.sutd.edu.sg, yuxizhou@whu.edu.cn} \\
{\tt\small jia\_gong@mymail.sutd.edu.sg, 
jun\_liu@sutd.edu.sg, tuzhigang@whu.edu.cn}
}

\begin{document}
\maketitle
\footnote{First Author and Second Author contribute equally to this work.}
\input{real_sec/0_abstract}
\input{real_sec/1_intro}

\input{real_sec/2_rela}

\input{real_sec/2.5_RL}
\input{real_sec/3_method}

\input{real_sec/4_experiments}

\input{real_sec/5_conclusion}
{
    \small
    \bibliographystyle{ieeenat_fullname}
    \bibliography{main}
}

\end{document}

%% file: preamble.tex
\usepackage[linesnumbered,ruled,vlined]{algorithm2e}
\usepackage[dvipsnames]{xcolor}


%% file: real_sec/0_abstract.tex
\begin{abstract}
Knowledge distillation (KD) enhances the performance of a student network by allowing it to learn the knowledge transferred from a teacher network incrementally. Existing methods dynamically adjust the temperature to enable the student network to adapt to the varying learning difficulties at different learning stages of KD. KD is a continuous process, but when adjusting the temperature, these methods consider only the immediate benefits of the operation in the current learning phase and fail to take into account its future returns. To address this issue, we formulate the adjustment of temperature as a sequential decision-making task and propose a method based on reinforcement learning, termed RLKD. Importantly, we design a novel state representation to enable the agent to make more informed action (\textit{i.e.}, instance temperature adjustment). To handle the problem of delayed rewards in our method due to the KD setting, we explore an instance reward calibration approach. In addition, we devise an efficient exploration strategy that enables the agent to learn valuable instance temperature adjustment policy more efficiently. Our framework can serve as a plug-and-play technique to be inserted into various KD methods easily, and we validate its effectiveness on both image classification and object detection tasks. 
Our code is at \href{https://itkd123.github.io/ITKD.github.io/}{ https://itkd123.github.io/ITKD.github.io/}. 
\end{abstract}

%% file: real_sec/1_intro.tex
\section{Introduction}
\label{sec:intro}

\begin{figure}[ht]
  \centering
   \includegraphics[width=0.67\linewidth]{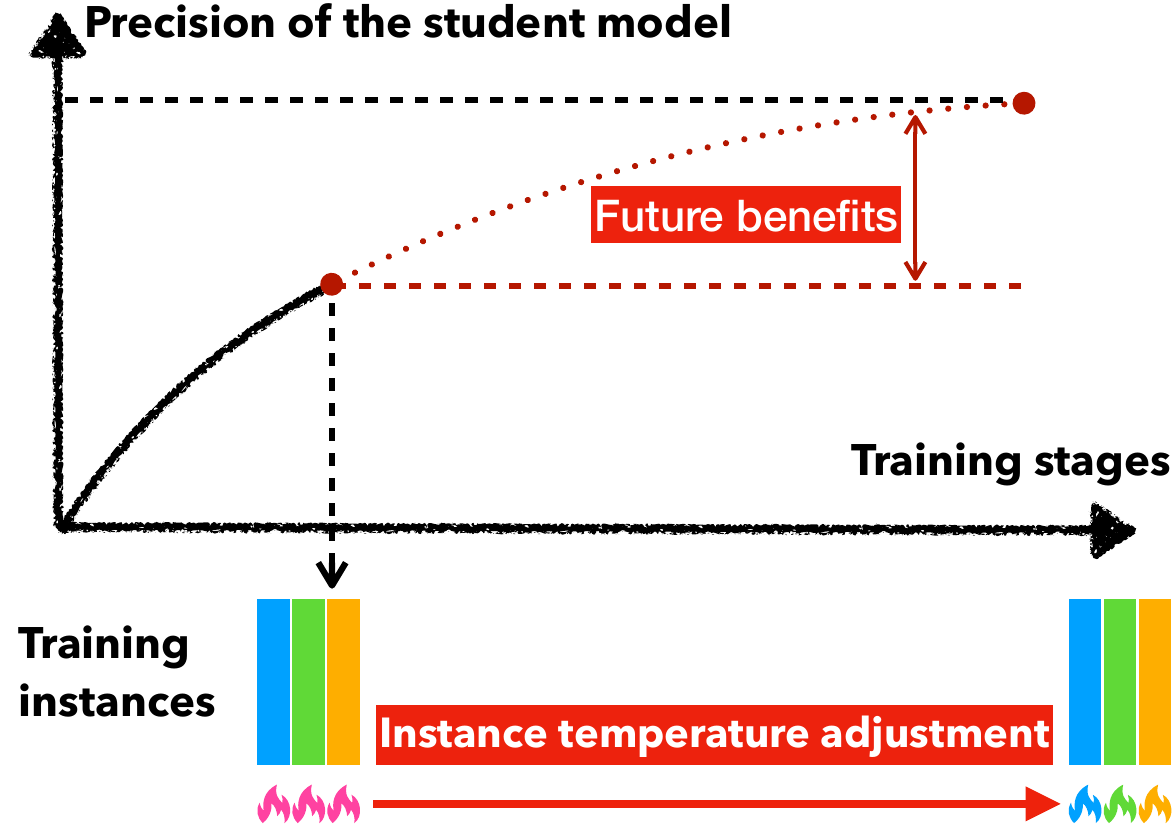}

   \caption{KD is a continual process, however, the previous KD methods~\cite{liu2022meta,li2023curriculum} do not consider the future benefits of instance temperature adjustment during the KD process.}
   \vspace{-2mm}
   \label{fig:motivation}
\end{figure}

Over the past few decades, the field of computer vision has undergone a transformative shift thanks to the remarkable progress of deep neural networks (DNNs). Nonetheless, the significant computation and storage demands of DNNs, present great challenges, especially in industrial applications where there is a preference for efficient and lightweight models. Typically, lightweight networks do not perform as well as deeper networks. To solve this issue, \textbf{knowledge distillation (KD)}, which aims to enable smaller (student) models to compete with larger (teacher) models in performance, has been introduced~\cite{hinton2015distilling}. Due to its remarkable effectiveness in boosting the capability of lightweight models, KD has been widely used in various tasks, \textit{e.g.}, object detection~\cite{chen2017learning,zhang2020improve}, semantic segmentation~\cite{liu2019structured,yang2020knowledge}, and natural language processing~\cite{tang2019natural,hahn2019self}.

KD enhances the student network by transferring knowledge from a higher-capacity teacher network. During the process of KD, the capability of the student network is constantly changing. This results in the same piece of knowledge (training instance) has varying degrees of value to the student network at different learning stages~\cite{li2022knowledge}. Moreover, even within the same learning stage, the difficulty of learning varies between instances. The student network should assign more weight to examples that are difficult to learn~\cite{kim2021self}. However, most previous KD methods~\cite{hinton2015distilling,zhao2022decoupled,yang2022cross,guo2023class} have not simultaneously taken into account the learning difficulty of each training instance as well as the learning stage they are in.

To address this issue, recent efforts~\cite{liu2022meta,li2023curriculum} have been made, where the temperature for each instance is adjusted to match its respective learning difficulty. This is because temperature, as a critical hyperparameter in KD, modulates the smoothness of the predictive distribution and sets the difficulty of the KD process. However, as illustrated in~\cref{fig:motivation} the previous methods~\cite{liu2022meta,li2023curriculum} do not taken into account the continuous nature of KD. When adjusting the instance temperature, it is important to consider the future benefits of this operation. Our key insight is that formulating the instance temperature adjustment in the KD process as a sequential decision-making task, with the adjustment of instance temperature being treated as the action in this task. For this sequential decision-making task, we set the reward as the performance improvement of the student network between two learning stages. Our goal is to maximize cumulative rewards, that is, to maximize the enhancement of the student network’s performance over the course of the task. To achieve this goal, we propose a novel method based on reinforcement learning (RL), termed as RLKD. 



In the proposed RLKD, we employ an agent network to determine the instance temperature. To aid the agent in making prudent decisions regarding instance temperature, we devise a comprehensive state representation which encompasses two performance features and an uncertainty feature. The two performance features respectively represent the teacher and student networks' performance on each instance, while the uncertainty feature is used to measure the student network's mastery over the instance. This innovative uncertainty feature design is arose by uncertainty-based sampling in active learning~\cite{roth2006margin,balcan2007margin}. Nevertheless,  we discover that applying the RL framework directly to instance temperature adjustment in KD can cause a significant delayed reward issue.

Given the setting of our reward (performance improvement of the student network) and the setup of KD, we calculate the reward at the end of training for each batch. Since a batch typically contains many training instances, often $32$, it means that the agent receives rewards only after the $32^{th}$ actions. This brings us a challenge of delayed rewards, causing difficulties in credit assignment~\cite{ke2018sparse,hernandez2018multiagent}. We explore an instance reward calibration method to handle this challenge, building upon the refinement of reward decomposition~\cite{arjona2019rudder}. Due to the absence of ground truth for the instance temperature, we adopt an online training mode to update the agent's policy.  However, in the initial phase of training, the agent may engage in random exploration within a vast and inefficient action space~\cite{amin2021survey}. To address this issue, we design an efficient exploration strategy that guides the agent to learn on high-quality training instances during the early phase of training. This strategy aims to expedite the agent's learning process, enabling it to quickly acquire valuable instance temperature adjustment policy.




In summary, our main contributions are:
\textbf{1)} In KD, to account for the future benefits of adjusting instance temperature at the current stage, we formulate the instance temperature adjustment as a sequential decision-making task, and propose a novel method RLKD based on RL to handle this task. \textbf{2)} To overcome the challenge of delayed rewards in our RLKD, we exploit a mechanism for instance reward calibration. Furthermore, we design a valid exploration strategy to promote the agent to learn valuable temperature adjustment policy with high efficiency. \textbf{3)} Our RLKD can serve as a plug-and-play technique to boost the performance of KD algorithms. We validate its effectiveness on three benchmarks for image classification and semantic segmentation, all obtaining the state-of-the-art result.

%% file: real_sec/2_rela.tex
\section{Related Work}
\label{sec:related_work}



KD, as a model compression method, can trace back its origin in ~\cite{hinton2015distilling}.  In the KD process, KL-divergence loss between teacher and student model predictions is minimized using a key hyperparameter known as ``temperature". As~\cite{hinton2015distilling, chandrasegaran2022revisiting,liu2022meta,li2023curriculum} stated, temperature helps adjusting the smoothness of the prediction distribution and sets the KD process's difficulty effectively. Due to the student model's learning capacity varies at different stages~\cite{li2022knowledge}, some works~\cite{liu2022meta,li2023curriculum} explore to adjust the temperature dynamically based on the current learning stage to help the student network learn better from the teacher network. 


MKD~\cite{liu2022meta} learns a dynamically varying temperature via the method of meta-learning~\cite{vilalta2002perspective} as the KD process progresses, but it is primarily designed for scenarios involving vision transformer~\cite{dosovitskiy2020image} and strong data augmentation. The limitations of MKD preclude its effective application in temperature adjustment within the majority of KD methods, and previous studies~\cite{li2023curriculum} have confirmed that directly applying MKD to KD models results in a significant degradation in performance. CTKD~\cite{li2023curriculum} utilizes a curriculum learning approach~\cite{bengio2009curriculum} to progressively learn a dynamic temperature parameter, starting from simple to complex scenarios. Particularly, CTKD progressively learns two versions of temperature: the global temperature and the instance temperature. However, CTKD does not take into account the future benefits (the performance enhancement of the student network between adjacent learning stages) when adjusting the instance temperature,  and it also does not consider the student network's mastery of the instance. These shortcomings of CTKD causing its instance temperature being not robust, preventing the student network trained with CTKD from learning knowledge effectively. To overcome these drawbacks, we formulate the instance temperature adjustment in the KD process as a sequential decision-making task, where adjusting instance temperature is considered as an action. Besides, we present a novel state representation, which includes a feature that reflects the student network's degree of mastery over a training instance.

%% file: real_sec/2.5_RL.tex
\section{Preliminary on Reinforcement Learning}

\begin{figure*}[ht]
  \centering
   \includegraphics[width=0.8\linewidth]{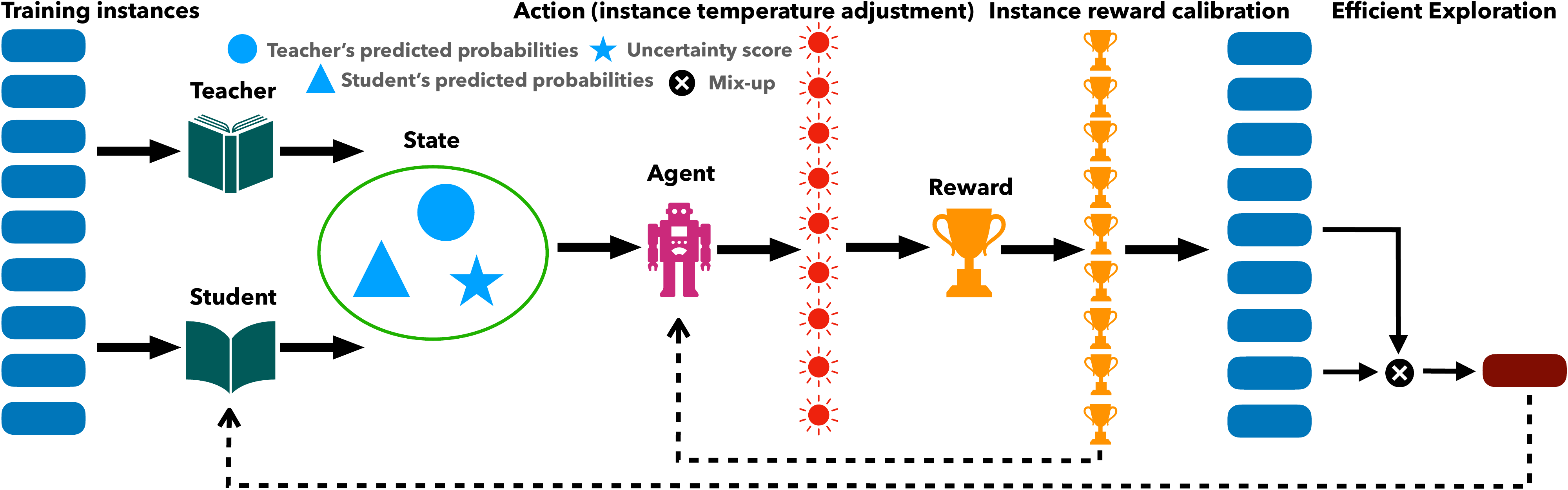}

   \caption{Overview of our RLKD method. Solid lines represent the processing flow of the training instances in our framework, and dashed lines indicate the backpropagation process used for model (student model and agent) updates. The workflow of our RLKD method is as follows: \textbf{1)} Given a batch of training instances, we first calculate the state through the outputs of the teacher and student networks, including the teacher's predicted probabilities, the student's predicted probabilities, and the student's uncertainty score. \textbf{2)} Our agent makes the decision (action) on the temperature for each training instance based on the state. \textbf{3)} Reward is calculated depending on the action that taken by the agent, followed by the instance reward calibration. \textbf{4)} According to the knowledge value of each training instance, we select the top 10-20\% highest-valued instances and the latter 40-50\% instances to perform a mix-up operation~\cite{zhang2017mixup}, accordingly high-quality training instances are obtained for the efficient exploration strategy.}
   \label{fig:pipeline}
\end{figure*}

Reinforcement Learning (RL) involves an agent aiming to gain maximum cumulative rewards through interactions with its environment. Key components include the Agent (the decision-maker), Environment (the context in which the agent operates), Action $A$ (choices made by the agent), State $S$ (the current situation), Reward $R$ (feedback from the environment), and Policy $\pi$ (the agent's strategy at a specific instance).

In RL, an agent, at state \(s_t\), decides on action \(a_t\) based on its policy, transitions to a new state \(s_{t+1}\), and receives reward \(r_{t+1}\). The agent's objective is to maximize its expected cumulative rewards and considering future gains, rather than just focusing on the immediate rewards. This is captured by the equation:
\begin{equation}
\vspace{-0.7mm}
 G_t = \sum_{k=0}^{\infty} \gamma^k r_{t+k+1} 
 \vspace{-0.4mm}
\end{equation}
Here, \( \gamma \) is the discount factor, lying between 0 and 1. It assigns the importance to future rewards, with values close to 1 implying a consideration for long-term rewards and values near 0 stressing on immediate rewards.

The Value Function \( V(s) \) for a policy \( \pi \) predicts the return from state \(s\):
\begin{equation}
 V^{\pi}(s) = \mathbb{E}_{\pi} [ G_t | S_t = s ] 
\end{equation}
The Q-Function \( Q(s, a) \) for a state-action pair \((s, a)\) with policy \( \pi \) is:
\begin{equation}
 Q^{\pi}(s, a) = \mathbb{E}_{\pi} [ G_t | S_t = s, A_t = a ],
\end{equation}
where \( \mathbb{E}_{\pi} \) represents the expectation under the policy \( \pi \). 
They relate via the Bellman Equation:
\begin{equation}
 V^{\pi}(s) = \sum_{a} \pi(a|s) \sum_{s', r} p(s', r|s, a) [ r + \gamma V^{\pi}(s') ].
\end{equation}

Our proposed RLKD method is based the Proximal Policy Optimization (PPO)~\cite{schulman2017proximal} framework. PPO, stemming from the policy gradient technique, tackles RL's issues of stability and efficiency. The actor in PPO optimizes the policy based on the feedback from the critic. The standard policy gradient techniques might make significant policy updates leading to erratic results, in contrast, PPO ensures the updates are restrained by using a clipped function.

The PPO objective is defined as:
\begin{equation}
 L^{CLIP}(\theta) = \hat{\mathbb{E}}_t [ \min(r_t(\theta) \hat{A}_t, \text{clip}(r_t(\theta), 1-\epsilon, 1+\epsilon) \hat{A}_t) ] 
\end{equation}
where
\( r_t(\theta) \) is the probability ratio of current to old policy action.
\( \hat{A}_t \) estimates the advantage function at time \( t \) where the critic provides a feedback.
The clip function keeps the ratio within a limited set by \( \epsilon \).

PPO's moderate updates ensure consistent learning, making it is suitable for our work, especially in the case that consistent learning is needed across varied settings.

%% file: real_sec/3_method.tex
\section{Method}

The previous KD methods~\cite{liu2022meta,li2023curriculum} attempt to adjust the temperature to improve the student network’s knowledge acquisition, but they overlook the nature of KD is continuous. When adjusting the temperature, they only consider the benefits in the current stage, neglecting the potential rewards of temperature adjustments in future learning stages. To address this issue, we treat the adjustment of instance temperature during the KD process as a sequential decision-making task , where the temperature adjustment for each instance is considered as the action within the task. Based on this insight, we propose the RLKD method (see~\cref{fig:pipeline}) based on RL with a novel state representation (described in \cref{subsection:RL}), allowing us to take into account the future rewards of temperature adjustment on training instances at the current stage. In our RLKD method, the reward is designed to measure the improvement in the student network's performance; thus, we calculate the reward during the parameter update of the student network. According to the KD setup, the student network updates its parameters after training on each batch of data (typically comprising 32 training instances), which means that we can only compute the reward once after every 32 actions. This leads to a significant delayed reward issue. To solve this problem, we design an instance reward calibration scheme (described in \cref{subsection:reward}). Furthermore, we formulate a strategy for efficient exploration, enabling the agent to rapidly learn effective temperature adjustment policy (described in \cref{subsection:high-quality}). 

\subsection{Instance Temperature Adjustment as a Sequential Decision-making Task}
\label{subsection:RL}

In this work, we aim at learning a policy that directly maximizes the performance of the student network driven by the maximization of our designed reward. To achieve this goal, we formulate the instance temperature adjustment in the KD process as a sequential decision-making task: $\left(s_t, a_t, r_{t+1}, s_{t+1}\right)$. Specifically, the process includes the following steps: \textbf{1)} Estimate the state $s_t$ based on the performance of the teacher and student networks on the current training instance. \textbf{2)} Given the current state $s_t$ and informed by the prior experiences, the agent evaluates each state-action pair and execute the action $a_t$ of temperature adjustment for each training instance. \textbf{3)} After the agent performs the optimized action $a_t$, the environment transfers to a subsequent state $s_{t+1}$ and provides a reward $r_{t+1}$ to the agent. \textbf{4)} The agent updates its policy based on the received reward $r_{t+1}$ and the newly observed state $s_{t+1}$. 

We utilize the PPO framework~\cite{schulman2017proximal} to model this process. Subsequently, we provide a detailed introduction to the definitions of state \( s_{t} \), action \( a_{t} \), and reward \( r_{t} \).

\textbf{State.} The state \( s_{t} \) serves as input for the agent, providing critical support for agent making the instance temperature decision. The design of the state should align with the needs of the instance temperature decision-making policy. Intuitively, when the policy makes a temperature decision for a training instance \( x \), it needs to consider the performance of both the teacher and the student networks on this instance. Moreover, due to the varying difficulty of the knowledge embodied in each training instance, the student network's mastery over each instance differs~\cite{xu2023computation,kim2021self}. The state \( s_{t} \) should also include a measure of the student network's grasp on that particular instance.

Based on these intuitions, given an instance \( x \), we collect cues from three aspects to form the state  \( s_{t} \): the performance of the teacher network, the performance of the student network, and the extent to which the student network has mastered the instance. Particularly, the teacher network outputs its prediction $p_{t}$ at instance \( x \) is expressed as:
\begin{equation}
p_{t}=\operatorname{argmax}_{i \in[k]} f_{{teacher }}(x)_i ,
\end{equation}
where, 
$k$ represents the total number of categories, and $f_{{teacher }}$ denotes the teacher network.
We use the probability $f_{{teacher }}(x)_{p_{t}}$ associated with the teacher network’s prediction $p_{t}$ for the instance $x$ to measure the performance of the teacher network. Similarly, to measure the performance of the student network, we use the probability $f_{{student }}(x)_{p_{s}}$ associated with the student network's prediction $p_{s}$ for that instance $x$. To assess the mastery level of a student network over the instance $x$, we draw inspiration from uncertainty-based sampling in active learning~\cite{roth2006margin,balcan2007margin}, and determine the mastery level by measuring the uncertainty score $\upsilon_{{student }}(x)$ in the student network’s prediction distribution for the instance. The uncertainty score for the student network with respect to instance $x$ is calculated according to:   
 \begin{equation}
\upsilon_{{student }}(x) = 1- (f_{{ {student }}}(x)_{p_{s}}-\max _{i \in[k] \backslash p_{s}} f_{{ {student }}}(x)_i).
\end{equation}
Our uncertainty score $\upsilon_{{student }}(x)$ is positively correlated with the degree of uncertainty exhibited by the student network towards instance $x$, which is because if the student network has a good grasp of the instance, the network is very confident in its prediction, resulting in a prediction distribution with a single high-probability predicted value. Conversely, if the mastery is poor, the student network exhibits uncertainty in its prediction, leading to multiple high-probability predicted values that are close to each other. 

In summary, for a given instance $x$, we define our state $s_t$ as $(f_{{teacher }}(x)_{p_{t}}, f_{{student }}(x)_{p_{s}}, \upsilon_{{student }}(x))$, encompassing the predicted probabilities from the teacher network, the predicted probabilities from the student network, and the uncertainty score of the student network. 


\textbf{Action.}
Our action is the decision-making regarding instance temperature $\mathcal{T}$. To overcome the limitations of exploration in a discrete action space, we opt to explore instance temperature $\mathcal{T}$ in a continuous action space. Below, we elaborate how to obtain the instance temperature $\mathcal{T}$. 

Upon receiving the state \( s_t \) for the instance $x$, we use \( s_t \)  as input to the actor network within the PPO framework. To better explore various actions of the actor network and to smooth its learning process, we design our actions to follow a Gaussian distribution $\mathcal{N}\left(\mu, \sigma^2\right)$. Thus our actor outputs the mean $\mu$ and variance $\sigma$ of a Gaussian distribution. To boost the flexibility and randomness of action exploration, we randomly sample a value from the Gaussian distribution to serve as our instance temperature $\mathcal{T}$. Finally, based on our experience that almost all the instance temperature varies within the range of 0 to 10, we restrict the temperature to a range by following formula:
\begin{equation}
\mathcal{T}=10\cdot\operatorname{sigmoid}(\mathcal{T}),
\end{equation}
where $\operatorname{sigmoid}$ refers to the sigmoid activation function.

\textbf{Reward.}
The reward function is a critical component of our framework, providing feedback regarding the quality of the agent's action, thereby assisting the agent in refining its action policy. The action of our agent is to select an appropriate instance temperature $\mathcal{T}$ based on the \( s_t \) of the instance \( x \), which can facilitate knowledge acquisition by the student network, aiming to maximize the performance of the student network as much as possible. To achieve this objective, we integrate the settings of KD and consider the improvement of the student network's performance between two consecutive batches as the reward. Moreover, a common characteristic in deep learning is that the student network's performance shows significant improvement during the initial stages of training, this may bring disproportionately large reward values. However, these large values do not necessarily reflect the agent's astute action choices. To mitigate the impact of this phenomenon, we progressively increase the reward size during the early training stages. The formula for the reward is defined as:
\begin{equation}
\label{eq:reward}
reward=\operatorname{sigmoid} (\mathcal{E}/n) \cdot reward.
\end{equation}
Herein, $\mathcal{E}$ represents the current epoch number, $n$ is a hyperparameter denotes the first $n$ epochs during which the reward incrementally grows.


\subsection{Instance Reward Calibration}
\label{subsection:reward}

In our RLKD method, the action is to adjust the temperature for each training instance. To evaluate the quality of a particular action, we should calculate the corresponding reward for that action. However, we are unable to directly obtain the instance reward for each action. This is because our reward is based on the performance improvement of the student network. The student model is trained on batches of instance data and updates its parameters accordingly. The reward can only be computed after the student model updates its parameters. Typically, in KD, the batch size is set to 32, meaning that we have to go through 32 actions before we can receive a reward. This delayed reward characteristic (known as the credit assignment problem~\cite{ke2018sparse,hernandez2018multiagent}) makes it is difficult to assess and improve the policy network. 

To address this issue, we design a reward corrector $\mathcal{C}$ based on the refinement of the reward decomposition~\cite{arjona2019rudder}. The reward corrector, which redistributes the reward $r^b$ for the current batch based on the state $s^{b}$ of the current batch and the action $a^{b}$ taken by the agent for each instance, to obtain the corrected reward $r^{b^\prime}$ that corresponds to the action for each instance. The corrected reward $r^{b^\prime}$ is calculated as:
  \begin{equation}
    \label{eq:instance reward}
        r^{b^\prime} = \mathcal{C} (s^{b}, a^{b}, r^b),
    \end{equation}
To account for the contribution of each instance's action to the reward of the entire batch, we introduce an auxiliary task that allows the reward corrector to predict the sequence-wide return $G$ at each time step. The loss function for our reward corrector $\mathcal{C}$ is defined as:
    \begin{equation}
        \mathcal{L}_\mathcal{C} = \alpha \cdot (r^{b^\prime}_n - r^b)^2 + \frac{\beta}{n} \cdot \sum_{i=1}^{n} (r^{b^\prime}_i - G_i)^2.
    \end{equation}
Here, \( r^{b^\prime}_n \) represents the \( n \)-th corrected reward, and \( n \) is the batch size. The variables \( a \) and \( b \) are weights, which we set to 1 and 0.5, respectively. The variable \( G_i \) denotes the return at the \( i \)-th time step. Additionally, to ensure that the states $s^{b}$ recorded in the replay buffer match the corrected rewards, we devise a state updater $\mathcal{U}$ and update the states $s^{b}$ accordingly. The updated state $s^{b^\prime}$ is calculated as follows:
    \begin{equation}
    \label{eq:update state}
      s^{b^\prime}   = \mathcal{U}(s^{b})
    \end{equation}
The loss function for our state updater $\mathcal{U}$  is defined as:
    \begin{equation}
        \mathcal{L}_\mathcal{U} = (\mathcal{E}(s^{b^\prime}) - G)^2.
    \end{equation}
Here, $\mathcal{E}$ refers to an estimator  to predict the corresponding return $G$ based on the input state.


\subsection{Efficient Exploration}
\label{subsection:high-quality}
Due to the lack of ground truth for instance temperature, the update process in our RL component is conducted online. In this training setup, it is imperative for the agent within the RL framework to quickly learn effective temperature adjustment policy. To enable our agent to adjust the temperature for each instance with higher accuracy, we set the action space as a continuous space. This often implies that, in the initial stages of training, the agent may engage in inefficient exploration across a vast action space~\cite{amin2021survey}, which is not conducive to rapidly learning valuable instance temperature adjustment policy. To solve this problem, we propose an efficient exploration strategy. In which, during the early stages of training the RL component, we guide the agent to learn on high-quality training instances, which is helpful to drive the agent towards more effective exploration.

Firstly, we need to define what constitutes high-quality training data in the context of KD. We consider that in KD, high-quality training samples mean those can provide more knowledge to the student network. The prior work~\cite{li2022knowledge} reveals the predictive entropy of a student network for a training instance can be used to measure the knowledge value of that instance. The higher the prediction entropy, the greater the knowledge value of the instance.
The prediction entropy of a student network for a training instance is:
\begin{equation}
H(y \mid x)=-\sum_{c=1}^C p(y=c \mid x) \log p(y=c\mid x).
\end{equation}
\begin{algorithm}[ht]
\caption{Training and usage of RLKD}
\label{alg:rlkd_training}
\SetAlgoLined
\SetNoFillComment
\SetKwInput{KwData}{Input}
\footnotesize
\KwIn{
agent network $ Q $, student network $f_S$, teacher network $f_T$, dataset $D$, batch size $N$, reward corrector $\mathcal{C}$
}
\For{batch $d_{t+1}$ in $D$}{
    \For{$i=0$ to $N-1$}{
        Build state observation $s_i$ from $f_S(d_{t+1}^{i})$, $f_T(d_{t+1}^{i})$ \\
        Compute the action $a_i$ and values $V_i \leftarrow Q(s_i)$ \\
        Use $a_i$ as instance temperature \\
        Collect $s_i$, $a_i$, $V_i$ as a buffer to update $Q$~(\cref{subsection:RL})
    }
    Update the student network $f_S$ \\
    Obtain the reward $r_{t+1}$ following~\cref{eq:reward}  \\
    Calibrate the instance reward following~\cref{eq:instance reward} \\
    Update the reward $r_{t+1}$ in the replay buffer \\
    Update the state $s_{t}$ following~\cref{eq:update state}~(\cref{subsection:reward})\\
    \While{not done}{
       Update $Q$ 
    }
}
    Compute prediction entropy $H_{t} \leftarrow  f_S(d_{t+1})$ \\
    Obtain high-quality samples following~\cref{eq:mixup}~(\cref{subsection:high-quality}) \\
   Implement the efficient exploration strategy \\
     Update $ Q $  \\
\end{algorithm}
Here, \( H(y \mid x) \) is the predictive entropy for a given instance \( x \), \( C \) is the number of classes, and \( p(y=c \mid x) \) is the predicted probability that instance \( x \) belongs to class \( c \). To identify the high-quality data, we first compute the prediction entropy \( H \) of the student model for all training instances, and then sort these prediction entropies in descending order to form a sequence \( S_e\). \( S_e\) is defined as:
\begin{equation}
    S_e  = \{I_1, ..., I_n\}.
\end{equation}
Here, $n$ represents the total number of training examples, and $I_n$ represents the training instance ranked at ranking $n$. 
To mitigate the risk of overfitting, we take the top $10\%$ to $20\%$ training samples in the sequence  \(S_e^{(10\sim20)\%} \) as our high-quality training samples for efficient exploration.

Secondly, since the student network is typically a small model, to prevent overfitting through our high-quality learning and to enhance the robustness of the student model, we utilize mix-up~\cite{zhang2017mixup} on our high-quality training samples with the training instances ranked from $40\%$ to $50\%$, denoted as \(S_e^{(40\sim50)\%} \), in the sequence \( S_e \). The sequence of training instance $S_e^{t}$ after mix-up is calculated as:
\begin{equation}
\label{eq:mixup}
\begin{aligned}
S_e^{t}=\lambda \cdot S_e^{(10\sim20)\%}+(1-\lambda) \cdot S_e^{(40\sim50)\%} 
\end{aligned}
\end{equation}
The parameter \( \lambda \) is set to ensure that the knowledge in $S_e^{t}$ predominantly comes from higher-ranked training instances (\textit{i.e.} higher quality training instances).

\subsection{Training and Usage of RLKD}
Given a dataset \( D \), a teacher network \( f_T \), and a student network \( f_S \), our RLKD proceeds as follows. At beginning, we calculate the instance temperature $\mathcal{T}$ for all training instances in the batch, organized by~\cref{subsection:RL}. We record the state \( s_t \), action \( a_t \), and value \( V_t \) of this batch into the replay buffer, which serves as a reference for the agent's subsequent decision-making. Next, we calibrate the instance reward and update the state, as described in~\cref{subsection:reward}. Finally, as demonstrated in~\cref{subsection:high-quality}, we filter out high-quality training samples based on the performance of each training instance during this training stage and execute our efficient exploration strategy by utilizing these high-quality training instances. The procedure is depicted in~\cref{alg:rlkd_training}.



        
    

%% file: real_sec/4_experiments.tex
\section{Experiments}

\begin{table*}[ht]
\centering
\scalebox{0.6}{
\begin{tabular}{cccccccccccc}
\toprule
Teacher & RN-56 & RN-110 & RN-110 & WRN-40-2 & WRN-40-2 & VGG-13 & WRN-40-2 & VGG-13 & RN-50 & RN-32$\times$4 & RN-32$\times$4 \\
Acc & 72.34 & 74.31 & 74.31 & 75.61 & 75.61 & 74.64 & 75.61 & 74.64 & 79.34 & 79.42 & 79.42  \\
\midrule
Student & RN-20 & RN-32 & RN-20 & WRN-16-2 & WRN-40-1 & VGG-8 & SN-V1 & MN-V2 & MN-V2 & SN-V1 & SN-V2 \\
Acc & 69.06 & 71.14 & 69.06 & 73.26 & 71.98 & 70.36 & 70.50 & 64.60 & 64.60 & 70.50 & 71.82 \\
\midrule
Vanilla KD & 70.66 & 73.08 & 70.66 & 74.92 & 73.54 & 72.98 & 74.83 & 67.37 & 67.35 & 74.07 & 74.45 \\
+CTKD & 71.11 & 73.47 & 71.08 & 75.40 & 73.97 & 73.48 & 75.70 & 68.42 & 68.51 & 74.52 & 75.26 \\
+Ours & 71.40 & 73.81 & 71.44 & 75.79 & 74.17 & 73.75 & 76.01 & 68.73 & 68.75 & 74.84 & 75.55 \\
\bottomrule
\end{tabular}
}
\caption{Student network top-1 accuracy on CIFAR-100. Testing the performance of Vanilla KD as well as Vanilla KD with the incorporation of instance temperature adjustment using CTKD and our RLKD, respectively.}
\label{tab:cifar vanilla KD}
\end{table*}

\begin{table*}[ht]
\centering
\scalebox{0.6}{
\begin{tabular}{ccc|ccc|ccc|ccc|ccc|ccc}
\toprule
         & Teacher & Student & Vanilla  KD     & +CTKD & +Ours & PKT    & +CTKD & +Ours & RKD    & +CTKD & +Ours & SRRL    & +CTKD & +Ours & DKD    & +CTKD & +Ours \\ \midrule
Top-1    & 73.96   & 70.26   & 70.83  & 71.28 & 71.39      & 70.92  & 71.31 &  71.53     & 70.94  & 71.13 & 71.37 & 71.01  & 71.25 &  71.38     & 71.13  & 71.47 & 71.62\\
Top-5    & 91.58   & 89.50   & 90.31  & 90.33 &   90.51    & 90.25  & 90.30 &  90.42     & 90.33  & 90.34 &  90.45  & 90.41  & 90.42 &  90.52  & 90.31  & 90.44 &   90.56   \\ \bottomrule
\end{tabular}
}
\caption{Top-1 and Top-5 accuracy on ImageNet with ResNet-34 as teacher and ResNet-18 as student.}
\label{tab:imagenet}
\end{table*}

\begin{table}[ht]
\centering
\scalebox{0.58}{
\begin{tabular}{cccccccc}
\toprule
Teacher & RN-56 & RN-110 & RN-110 & WRN-40-2 & WRN-40-2 & RN-32$\times$4 & RN-32$\times$4 \\ 
Acc & 72.34 & 74.31 & 74.31 & 75.61 & 75.61 & 79.42 & 79.42 \\ 
\midrule
Student & RN-20 & RN-32 & RN-20 & WRN-16-2 & WRN-40-1  & SN-V1 & SN-V2 \\
Acc & 69.06 & 71.14 & 69.06 & 73.26 & 71.98 & 70.70 & 71.82  \\
\midrule
PKT   & 70.85 & 73.36 & 70.88 & 74.82 & 74.01 & 74.39& 75.10\\
+CTKD & 71.13 & 73.49 & 71.07 & 75.34 & 74.11 & 74.63 & 75.52 \\
+Ours & 71.41 & 73.68 & 71.34 & 75.62 & 74.23 &  74.89 & 75.78 \\
\midrule
SP  & 70.84 & 73.09 & 70.74 & 74.88 & 73.77 & 74.97 &75.59 \\
+CTKD & 71.29 & 73.42 & 71.17 & 75.30 & 73.97 & 75.28 & 75.79 \\
+Ours & 71.65 & 73.70 & 71.51 & 75.61 & 74.22 & 75.31 &  76.04\\
\midrule
VID &70.62 & 73.02 & 70.59 & 74.89 & 73.60 & 74.81 & 75.24  \\
+CTKD & 70.81 & 73.38 & 71.11 & 75.20 & 73.75 & 75.23 & 75.48 \\
+Ours & 71.09 & 73.70 & 71.39 & 75.48 & 74.02 & 75.58 & 75.81\\
\midrule
CRD & 71.69  & 73.63 & 71.38 & 75.53 & 74.36 & 75.13 & 75.90  \\
+CTKD & 72.13  & 74.08 & 72.02 & 75.71 & 74.72 & 75.41 & 76.20  \\
+Ours & 72.29 & 74.41 & 72.28 & 76.03 & 74.98 & 75.68 & 76.55 \\
\midrule
SRRL & 71.13  & 73.48 & 71.09 & 75.69 & 74.18 & 75.36 & 75.90 \\
+CTKD & 71.41 & 73.81 & 71.52 & 75.90 & 74.38 & 75.62 & 75.97 \\
+Ours & 71.61 & 74.02 & 71.81 & 76.23 & 74.64 & 75.90 & 76.06\\
\midrule
DKD & 71.43 & 73.66 & 71.28 & 75.70 & 74.54 & 75.44 & 76.48 \\
+CTKD & 71.62 & 73.91 & 71.65 & 75.85 & 74.57 & 75.88 & 76.91 \\
+Ours & 71.89 & 74.27 & 71.91 & 76.02 & 74.90 & 76.02 & 77.21 \\
\bottomrule
\end{tabular}
}
\caption{Student network Top-1 accuracy on CIFAR-100 dataset. }
\vspace{-5mm}
\label{tab:cifar various KD}
\end{table}

For a fair comparison, we follow the experimental settings of CTKD~\cite{li2023curriculum} to conduct experiments to verify the effectiveness of our RLKD. Experiments are tested on a variety of well-known neural network architectures, such as VGG~\cite{simonyan2014very}, ResNet (RN)~\cite{he2016deep}, Wide ResNet (WRN)~\cite{zagoruyko2016wide}, ShuffleNet (SN)~\cite{zhang2018shufflenet}, and MobileNet (MN)~\cite{howard2017mobilenets}. We also evaluate RLKD as a plug-and-play technique across various distillation frameworks, including Vanilla KD~\cite{hinton2015distilling}, PKT~\cite{passalis2018learning}, SP~\cite{tung2019similarity}, VID~\cite{ahn2019variational}, CRD~\cite{tian2019contrastive}, SRRL~\cite{yang2020knowledge}, and DKD~\cite{zhao2022decoupled}. Furthermore, we perform ablation studies to validate the effectiveness of our designed state representation, instance reward calibration, efficient exploration strategy, and selection of high-quality training examples.

\textbf{Tasks and datasets.} Following~\cite{li2023curriculum}, we conduct experiments on two tasks: image classification and object detection. For the image classification task, we carry out experiments on CIFAR-100~\cite{krizhevsky2009learning} and ImageNet~\cite{deng2009imagenet}. For the object detection task, we conduct our experiments on MS-COCO~\cite{lin2014microsoft}. CIFAR-100 is a prominent dataset for image classification, comprising 32$\times$32 pixel images across 100 different categories, with a training set of 50,000 images and a validation set of 10,000 images. ImageNet, another significant dataset for large-scale image classification, encompasses 1,000 categories with a training set of approximately 1.28 million images and a validation set of 50,000 images. MS-COCO is a famous dataset used for general object detection that includes 80 categories. It has a training set (train 2017) with 118,000 images and a validation set (val 2017) with 5,000 images.

\subsection{Main results}

\textbf{CIFAR-100: image classification.} As shown in~\cref{tab:cifar vanilla KD}, we conduct image classification on the CIFAR-100 dataset to demonstrate the generalization performance of our RLKD method across 11 teacher-student pairs, including RN-56 \& RN-20, \textit{etc.} Among them, 5 pairs of teacher and student models (VGG-13 \& MN-V2, \textit{etc.}) are characterized by distinguishing architectural frameworks. These experimental designs we employed provide a diverse and comprehensive assessment environment. When the teacher and student networks share the same architecture, the experimental results show that our RLKD method has a strong generalization capacity, also exhibits a superior performance compared to CTKD. Specifically, in the case of RN-110 \& RN-20, our method outperforms Vanilla KD by 0.78\% (71.44\% vs 70.66\%)  and CTKD by 0.36\% (71.44\% vs 71.08\%). Moreover, in the case where the teacher and student networks have different architectures, the powerful generalization capacity of our RLKD is also validated. 

To validate the generalization of our RLKD method across different KD frameworks, we conduct experiments on 6 currently leading KD frameworks (see~\cref{tab:cifar various KD}), including DKD, PKT, \textit{etc.} When applied to the teacher-student pair RN110 \& RN32, our RLKD brings an improvement of 0.61\% (74.27\% vs 73.66\%) in the DKD framework, which surpasses the accuracy of CTKD by 0.36\% (74.27\% vs 73.91\%). Experiments conducted on other 5 KD frameworks (\textit{e.g.} PKT, \textit{etc.}) further confirm the strong generalization of our RLKD. Both the accuracy and stability of the proposed RLKD are significantly superior to CTKD, this can be attributed to our RLKD method considers the future rewards of the instance temperature adjustment operations.

\begin{table}[ht]
\centering
\scalebox{0.63}{
\begin{tabular}{ccccccc}
\toprule
        & mAP   & AP50  & AP75  & APl   & APm   & APs   \\ \midrule
T: RN-101 & 42.04 & 62.48 & 45.88 & 54.60 & 45.55 & 25.22 \\
S: RN-18  & 33.26 & 53.61 & 35.26 & 43.16 & 35.68 & 18.96 \\ \midrule
Vanilla KD& 33.97 & 54.66 & 36.62 & 44.14 & 36.67 & 18.71 \\
+CTKD     & 34.51 & 55.32 & 36.95 & 44.76 & 37.17 & 19.01 \\ 
+Ours     & 34.73 & 55.61 & 37.19 & 45.27 & 37.30 & 19.12 \\ \midrule \midrule
T: RN-50  & 40.22 & 61.02 & 43.81 & 51.98 & 43.53 & 24.16 \\
S: MN-V2  & 29.47 & 48.87 & 30.90 & 38.86 & 30.77 & 16.33 \\ \midrule
Vanilla KD& 30.13 & 50.28 & 31.35 & 39.56 & 31.91 & 16.69 \\
+CTKD     & 31.21 & 52.12 & 32.01 & 41.11& 33.44 & 18.09 \\ 
+Ours     & 31.49 & 52.57 & 33.23 & 41.71 & 33.65 & 18.31 \\ \bottomrule
\end{tabular}
}
\caption{Results of our RLKD on the MS-COCO dataset, utilizing Faster-RCNN~\cite{ren2015faster} with FPN~\cite{lin2017feature}. We conduct experiments with the following teacher-student pairings: RN-101 paired with RN-18,  and RN-50 paired with MN-V2.}
\vspace{-5mm}
\label{tab:ms-coco}
\end{table}

\textbf{ImageNet: image classification.} To validate the scalability of our method and its applicability in complex scenarios involving large datasets, we further conduct image classification on ImageNet.~\cref{tab:imagenet} details the top-1 and top-5 accuracy. Using CTKD and our RLKD as the adaptable plug-in approach, we incorporate them into 5 current leading distillation frameworks (\textit{i.e.} KD, PKT, RKD, SRRL, and DKD). The experimental results obtained from these 5 KD frameworks unequivocally demonstrate the excellent scalability of our method. Remarkably, our RLKD exhibits robust performance on large dataset like ImageNet. For instance, in the Vanilla KD and SRRL frameworks, our method achieves improvement of 0.2\% (90.51\% vs 90.31\%) and 0.11\% (90.52\% vs 90.41\%) respectively. In contrast, CTKD obtains much fewer improvement on these KD frameworks, with gains of just 0.02\% (90.33\% vs 90.31\%) and 0.01\% (90.42\% vs 90.41\%) respectively, \textbf{about 10 times lower}. We think the superior performance of RLKD can be attributed to its RL-based framework in instance temperature adjustment, which considers the future benefits of these adjustments. Additionally, unlike CTKD, our RLKD also takes into account the student model's grasp of individual instances during instance temperature adjustment.




\textbf{MS-COCO: object detection.} To verify whether our RLKD method possesses robustness across other visual tasks, we execute object detection on the MS-COCO dataset. As shown in~\cref{tab:ms-coco}, in the case of RN-50 \& MN-V2, regarding the mAP metric, our RLKD outperforms Vanilla KD by 1.36\% (31.49\% vs 30.13\%) and CTKD by 0.28\% (31.49\% vs 31.21\%), respectively. Additionally, for detecting objects with varying sizes -- evaluated by the AP metrics for large (APl), medium (APm) and small (APs) objects, our RLKD also shows a significant enhancement, consistently surpasses CTKD across all size categories. Results demonstrate the robustness of our approach, where instance temperature adjustment is treated as a sequential decision-making task, enabling consideration of future benefits.




\begin{table}[ht]
\centering
\scalebox{0.66}{
\begin{tabular}{ccccccc}
\toprule
Teacher  & RN-56 & RN-110 & WRN-40-2 & VGG-13 \\
Student     & RN-20 & RN-32  & WRN-16-2 & VGG-8  \\
\midrule

 Ours w/o US  & 71.16     & 73.68      & 75.61    & 73.57  \\
 Ours w US & 71.40     & 73.81      & 75.79    & 73.75  \\
\bottomrule
\end{tabular}
}
\caption{Ablation study of the uncertainty score (US) feature. }
\label{tab:us}
\end{table}

\begin{table}[ht]
\centering
\scalebox{0.66}{
\begin{tabular}{ccccccc}
\toprule
Teacher  & RN-56 & RN-110 & WRN-40-2 & VGG-13 \\
Student      & RN-20 & RN-32  & WRN-16-2 & VGG-8  \\
\midrule

Ours w/o IRA  & 70.91     & 73.26      & 75.39    & 73.32  \\  
Ours w IRA & 71.40     & 73.81      & 75.79    & 73.75  \\
\bottomrule
\end{tabular}
}
\caption{Ablation on instance reward calibration (IRA) strategy.}
\label{tab:ira}
\end{table}

\begin{table}[ht]
\centering
\scalebox{0.66}{
\begin{tabular}{ccccccc}
\toprule
Teacher & RN-56 & RN-110 & WRN-40-2 & VGG-13 \\
Student       & RN-20 & RN-32  & WRN-16-2 & VGG-8  \\
\midrule

Ours w/o EE   & 71.03     & 73.52      & 75.50    & 73.45  \\
Ours w EE & 71.40     & 73.81      & 75.79    & 73.75  \\
\bottomrule
\end{tabular}
}
\caption{Ablation study of the efficient exploration (EE) strategy.}
\label{tab:HL}
\end{table}

\subsection{Ablation studies}
In the ablation studies, we evaluate the performance of the uncertainty score that is included in our state representation, the instance reward calibration scheme, the efficient exploration strategy, and different high-quality training example selection strategies. All experiments are conducted on the CIFAR-100 dataset with respect to the image classification task, and utilize the Vanilla KD framework.


\textbf{Uncertainty score.} We conduct experiments on 4 sets of teacher-student network pairs to test the effectiveness of the uncertainty score in our state representation. As shown in~\cref{tab:us}, when incorporating uncertainty score into state representation, our method shows an improvement of 0.24\% (71.40\% vs 71.16\%) in the RN-56 \& RN-20 teacher-student pair. This enhancement verifies the effectiveness of our designed uncertainty score, which enables the agent to make wiser decisions by taking into account the student model's mastery of the training instances.


\textbf{Instance reward calibration.} As shown in~\cref{tab:ira}, when incorporating an instance reward calibration strategy into our RLKD method, a promotive effect across 4 different sets of the teacher-student pairs (RN-56 \& RN-20, \textit{etc.}) is achieved. \textit{E.g.},  our instance temperature calibration strategy boosts the performance of RN-110 \& RN-32 pair by 0.55\% (73.81\% vs 73.26\%). We believe the effectiveness of the instance reward calibration strategy lies in its ability to enable the agent to more accurately perceive the rewards resulting from each of its instance temperature adjustment actions, thereby enhancing its capacity to update its policy for performing the action.

\textbf{Efficient exploration.} As shown in~\cref{tab:HL}, we conduct ablation experiments on our efficient exploration strategy across 4 teacher-student pairs. The experimental results  demonstrate that our effective exploration strategy\begin{table}[ht]
\centering
\scalebox{0.57}{
\begin{tabular}{cc|cccc}
\toprule
         Teacher & Student & $0\sim10\%$     & $10\sim20\%$ & $10\sim20\%$ $\mathcal{M}$ $30\sim40\%$ & $10\sim20\%$ $\mathcal{M}$ $40\sim50\%$     \\ \midrule
   72.34    & 69.06   & 70.92  & 71.21       & 71.27    & 71.40 \\
    75.61   & 73.26   & 75.33  & 75.57       & 75.61    &  75.79      \\ \bottomrule
\end{tabular}
}
\caption{Comparison of different high-quality training sample selection strategies. The teacher-student pairs corresponding to the second and third rows are respectively RN-56 \& RN-20 and WRN-40-2 \& WRN-16-2. ``$\mathcal{M}$'' denotes the mix-up operation.}
\label{tab:selection}
\end{table}
 facilitates performance of the student model across 4 teacher-student pairs. In the experiments involving the RN-56 \& RN-20 teacher-student pair, our efficient exploration strategy results in a performance improvement of 0.37\% (71.40\% vs 71.03\%). We attribute this success to the strategy enables the agent to learn valuable instance temperature adjustment policy faster, allowing the student model to acquire more useful knowledge during the early stages of KD.


\textbf{Selection of high-quality training examples.}  As shown in~\cref{tab:selection}, we conduct experiments on CIFAR-100 to compare different strategies for selecting the high-quality training examples. Interestingly, we observe that when using the top 10\% of high-quality training data, the performance of the student model in the teacher-student pair RN-56 \& RN-20 is 70.92\%, which is not as good as the performance 71.21\% of the student model when using the training data ranked from 10\% to 20\%. This phenomenon is also observed in the teacher-student pair WRN-40-2 \& WRN-16-2. We think this may due to utilizing the top 10\% samples caused overfitting in the agent. Furthermore, in the teacher-student pair RN-56 \& RN-20, when conducting the mix-up method on the training data ranked from 10\% to 20\% using the training data ranked 40\% to 50\%, there is a performance increase of 0.19\% (71.40\% vs 71.21\%). The experimental results verify the validity of our mix-up method that combines instances of varying knowledge values can produce high-quality training data.   

%% file: real_sec/5_conclusion.tex
\section{Conclusion}

In current knowledge distillation domain, the methods~\cite{li2023curriculum,liu2022meta} applied to temperature adjustment neglect the consideration of future benefits associated with the adjustment. To address this issue, we approach the instance temperature adjustment as a sequential decision-making task and propose a novel method RLKD. Specifically, we design a comprehensive state representation to enable the agent in our framework to make informed adjustment to the instance temperature. Besides, we explore an instance reward calibration scheme to provide the agent with more accurate reward signals. In addition, we develop an efficient exploration strategy to boost the agent's capability to learn valuable temperature adjustment policy fastly. Extensive experiments are conducted on three famous datasets for the tasks of image classification and object detection, demonstrating the effectiveness of our plug-and-play instance temperature adjustment method RLKD.